\documentclass[12pt]{article}
\usepackage{authblk}
\usepackage{graphicx}
\usepackage[utf8]{inputenc}
\usepackage{linguex}
\usepackage{booktabs} 
\usepackage{tabularx} 
\usepackage[margin=1in]{geometry} 
\usepackage{natbib}
\usepackage{setspace}
\title{Large Language Models as Modal Models in Linguistics}

\author[1]{Haruto Suzuki\thanks{\texttt{s17599@keio.jp}}}
\author[2,3]{Saku Sugawara \thanks{\texttt{saku@nii.ac.jp}}}

\affil[1]{Keio University}
\affil[2]{National Institute of Informatics}
\affil[3]{University of Tokyo}

\date{June 2026}

\begin{document}

\maketitle

\begin{abstract}

The rapid advancement of large language models (LLMs) has intensified debates about their significance for linguistic theory. These debates are commonly divided into three positions: insulationism, which regards LLMs as irrelevant to human language; eliminativism, which claims that LLMs can replace traditional linguistic theories; and conciliationism, which views them as useful tools for linguistic research. To clarify these positions, this paper applies the framework of modal modeling from the philosophy of science. We argue that LLMs possess genuine epistemic value as minimal models, even without structural correspondence to human cognition. In particular, they can provide \textit{how-possibly explanations} (HPEs) by testing modal claims about language acquisition and linguistic competence. We then examine the conditions under which LLMs could qualify as \textit{how-actually explanations} (HAEs) of human language, drawing on the mechanistic account of scientific explanation. We argue that current LLMs do not yet satisfy these requirements. On the basis of this analysis, we propose understanding the explanatory power of LLMs as lying on a continuum between HPEs and HAEs. This framework avoids both overstating and understating their explanatory significance and offers a more precise basis for evaluating the role of LLMs in the scientific study of language. 

\end{abstract}

\section{Introduction}

 Recent years have seen large language models (LLMs) such as GPT achieve remarkable performance across a wide range of natural language tasks, including machine translation, text summarization, and question answering. These developments have prompted renewed debate in linguistics about how human language should be explained and understood. Current discussions on the relationship between LLMs and linguistic theory can be broadly divided into three positions: The first is \textit{insulationism}, which views LLMs as irrelevant because they require an incomparably vast amount of data compared to humans and merely provide accurate predictions without offering scientific explanations (Kodner et al., 2023 \cite{kodner2023linguisticsthrive21stcentury}). The second is \textit{eliminativism}, which suggests that LLMs can replace traditional theories by acquiring syntax without explicit rules (Piantadosi, 2024 \cite{piantadosi2023modern}) \footnote{The labels ``insulationism'' and ``eliminativism'' for these two positions are adopted from McGrath et al. (2024) \cite{doi:10.1177/09637214241268098}.}. The third is what we call conciliationism, which treats LLMs as useful tools for testing and refining existing theories (Millière, 2024 \cite{milliere}; Futrell and Mahowald, 2025 \cite{Futrell_2025}). 

These debates are often rooted in a lack of shared understanding regarding meta-scientific concepts such as what counts as a scientific explanation and what scientific models are intended to achieve. To evaluate the validity of claims about the significance of LLMs in linguistics, it is necessary to make explicit the underlying meanings and assumptions that each position relies on. This paper evaluates the epistemic value of LLMs in linguistics by drawing on tools from the philosophy of science to analyze their relationship with linguistic theory. 
 
We propose to understand this relationship through the framework of \textit{modal modeling} (cf. Wirling and Grüne-Yanoff, 2021 \cite{S&G2011}). This approach involves modeling practices that aim to provide information about what is possible or what is necessary rather than merely describing the actual world. Within this framework, we distinguish between two types of explanations: how-possibly explanations (HPEs), which show how a phenomenon could occur in principle, and how-actually explanations (HAEs), which identify the actual mechanisms that cause a phenomenon (Bokulich, 2014 \cite{Bokulich2014-BOKHTT}).

First, we argue that LLMs possess genuine epistemic value even when considered as minimal models (Grüne-Yanoff, 2009 \cite{Grune-Yanoff2009-GRNLFM-2}) that lack structural isomorphism with human cognitive mechanisms. As minimal models, LLMs provide HPEs that can test modal claims in linguistics, such as the claim that language acquisition is impossible without innate language-specific constraints. This approach allows researchers to update their degree of confidence in beliefs about what is necessary or impossible regarding the architecture and acquisition of human language. Millière (2024) argues that language models trained under designs that consider cognitive and developmental plausibility can provide HPEs regarding language acquisition and linguistic competence \cite{milliere}. We expand on this view by applying the framework of modal modeling to clarify what kinds of beliefs in linguistics are updated by these explanations. Second, we examine the conditions required for LLMs to surpass HPEs and provide HAEs for human linguistic competence. Drawing on the mechanistic account of explanation (Craver, 2006 \cite{Craver2006-CRAWMM}; Kaplan and Craver, 2011 \cite{Kaplan}), we argue that LLMs must satisfy two requirements derived from the 3M++ constraint (Cao and Yamins, 2024 \cite{CAO2024101244}): \textit{Predictively Adequate Runnable Abstraction (PARA)}, which demands that models process the same type of input as humans and broadly predict human behavioral patterns; and \textit{Transform Similarity}, which requires a meaningful structural correspondence between model internals and human neural mechanisms at an appropriate abstraction. We show that current LLMs fall short of these requirements.

 \begin{figure}
    \centering
    \includegraphics[width=0.9\linewidth]{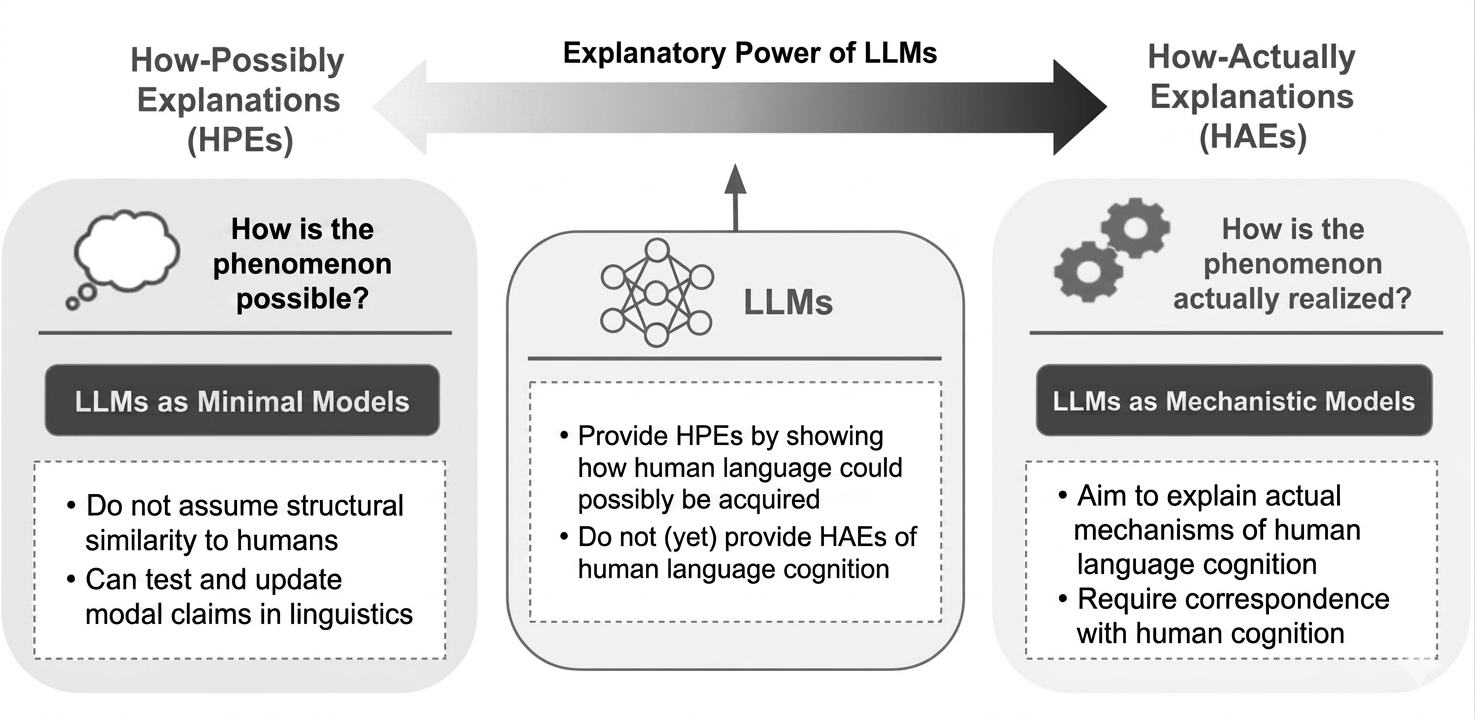}
    \caption{A framework for evaluating the explanatory power of LLMs
}
    \label{fig:placeholder}
\end{figure}

This analysis leads us to propose a continuum view of the explanatory power of LLMs, ranging from HPEs to HAEs (Figure 1). From this perspective, both insulationists and eliminativists share the same mistake of evaluating LLMs' explanatory power on an all-or-nothing basis: the former deny LLMs' value as HPEs simply because they fall short of HAEs, while the latter mistake HPE-level success for the establishment of HAEs. Furthermore, while the conciliationist position points in the right direction, its criteria for evaluating explanatory power remain unclear. Our framework addresses these issues and provides a common ground for more constructive discussions.

\section{Background}

\subsection{Existing Positions on the Relationship Between LLMs and Linguistics}

The rapid development of LLMs, exemplified by systems such as GPT, has renewed interest in a longstanding question in linguistics: can artificial systems that exhibit sophisticated linguistic behavior contribute to our understanding of human language? In response, three broad positions have emerged in the recent literature: insulationism, eliminativism, and conciliationism. Here, we examine each position in greater detail, paying particular attention to the methodological assumptions they rely on.

\textbf{Insulationists} maintain that the impressive linguistic performance of LLMs is irrelevant to theoretical linguistics (Chomsky et al. 2023 \cite{Chomskyetal2023}; Kodner et al. 2023 \cite{kodner2023linguisticsthrive21stcentury}; Fox and Katzir 2024 \cite{FoxandKazir}; Muller 2025 \cite{articleMuller}; Bolhuis 2025 \cite{bolhuis2025largelanguagemodelsnatural}). Their arguments typically rely on two considerations. The first concerns the gap between LLMs and human linguistic competence or language acquisition. For example, current LLMs rely on massive amounts of data, up to trillions of tokens, which is thousands of times more than the input human children are typically exposed to (Warstadt et al. 2023a  \cite{warstadt2023papersbabylmchallenge}). Additionally, while human infants quickly learn languages that follow natural language structures but struggle with impossible languages that have unnatural structures, language models easily learn these impossible languages just as well as real ones (cf. Moro et al. 2023 \cite{MORO202382}). Critics further argue that LLMs fail to capture central properties traditionally associated with human linguistic cognition, including compositionality, the competence–performance distinction, and modularity (e.g., Fox and Katzir 2024, Duple 2021 \cite{Dupre2021-DUPWCD}). The second consideration concerns the nature of scientific explanation itself. For insulationists, LLMs are essentially devices for generating probabilistic predictions. Even granting their high predictive accuracy, this alone does not meet the central role of scientific theory. As Kodner et al. (2023) argue, scientific theories are expected to provide interpretable and explanatory accounts, not merely accurate predictions. In this respect, LLMs resemble the Ptolemaic model of astronomy, which achieved high predictive accuracy through successive epicycles yet offered no genuine explanation of planetary motion (ibid. Sec.4). The underlying assumption of this position is that genuine scientific models or theories must reveal actual causal mechanisms and maintain structural correspondence with their target system. From this view, any model that departs from the known features of language acquisition and linguistic cognition is simply not a model that explains human language.

\textbf{Eliminativists} adopt a markedly different stance. They argue that the success of LLMs calls into question the necessity of many traditional linguistic theories and may even render them obsolete. Piantadosi (2024), for example, contends that the achievements of modern language models constitute a practical challenge to core assumptions of the Chomskyan tradition. In particular, LLMs acquire sophisticated syntactic and semantic regularities through predictive learning alone, without relying on explicitly encoded grammatical rules or an innate Universal Grammar. If such learning is possible, then the poverty of the stimulus argument appears substantially weakened \cite{piantadosi2023modern}. Some eliminativists go further and suggest that LLMs should themselves be regarded as superior linguistic theories. Ambridge and Blything (2024), for instance, argue that LLMs outperform traditional theories in predicting acceptability judgments related to verb argument structure \cite{A&B2024}. On this view, language resembles other complex adaptive systems, such as financial markets or social insect colonies, whose behavior emerges from large-scale interactions rather than from simple underlying principles. Consequently, linguistic theories may be inherently complex and resistant to intuitive understanding. Elegance and interpretability, traditionally regarded as theoretical virtues, may therefore be liabilities rather than strengths. The central assumption underlying eliminativism is that sufficiently accurate simulation of a system’s input–output behavior is enough to establish explanatory superiority. If an LLM predicts linguistic behavior better than competing theories, then it should be regarded as the better theory.

A third position, which we call \textbf{conciliationists}, seeks a middle ground between these extremes. They argue that LLMs neither replace existing linguistic theories nor remain irrelevant to them. Instead, they function as valuable tools for evaluating, refining, and constraining hypotheses about language acquisition and linguistic competence. Millière (2024) proposes that by carefully controlling learning conditions and making use of causal intervention, experiments with LLMs can potentially constrain hypotheses about language acquisition and competence \cite{milliere}. Futrell and Mahowald (2025) similarly argue that LLMs provide a proof of concept for formalizing and testing the gradient and probabilistic view of language advocated by usage-based models and functional linguistics \cite{Futrell_2025}. The assumption underlying this position is that scientific models do not need to share a strict structural correspondence with the human brain to offer valid linguistic explanations. Even highly idealized artificial systems may provide insight into human cognition, provided that they allow researchers to investigate relevant hypotheses under controlled conditions.

Despite their disagreements, all three positions share a common failure: they do not make explicit the meta-scientific concepts on which their arguments depend. When insulationists claim that LLMs \textit{predict} but do not \textit{explain}, and eliminativists claim they do \textit{explain}, it is unclear whether they are working with the same notion of explanation. When conciliationists speak of LLMs enabling reasoning about human language acquisition as models of language, the notion of \textit{model} they invoke is not shared across the debate. Without clarifying these underlying commitments, the conflict is in part merely verbal. To evaluate the epistemic value of LLMs fairly, it is therefore necessary to make these meta-scientific premises explicit. The philosophy of science offers precisely the tools needed for this task, and it is to these that we now turn.

\subsection{Modal Modeling}

In this paper, we develop a framework for evaluating the epistemic value of LLMs by drawing on recent discussions of \textit{modal modeling} in the philosophy of science. Modal modeling refers to modeling practices that aim to provide modal information, such as what is possible, what is necessary, or what could happen under counterfactual situations, rather than providing information about the actual world (cf. Wirling and Grüne-Yanoff 2025  \cite{SjolinWirling2025-SJLTTP}). Scientific inquiry frequently relies on such models. For example, researchers use models to investigate how much global temperatures could rise under particular conditions, what level of vaccination coverage would be required to contain a pandemic, or how a proposed housing policy might alter patterns of urban population distribution.

Traditional accounts of scientific modeling have generally assumed that models function by representing real-world target systems (e.g., Giere 2004  \cite{Giere2004-GIEHMA}; Weisberg 2013 \cite{Weisberg2013-WEISAS-5}). On this view, the primary aim of a model is to capture relevant features of actual entities, processes, or events. Consequently, highly idealized models that fail to accurately represent their targets have often been regarded as merely heuristic devices or as provisional steps toward more complete explanations (e.g., Craver 2006  \cite{Craver2006-CRAWMM}). Recent work in the philosophy of science, however, has challenged this assumption. In discussions of modal modeling, philosophers increasingly argue that a model can possess epistemic value even when it does not accurately represent its target system (cf. Knuuttila et al. 2025; Introduction  \cite{book})\footnote{A similar trend is observed in discussions regarding the epistemic status of machine learning models in general. Sullivan (forthcoming) argues that even when machine learning (ML) models do not appear to represent their targets from a representationalist perspective, they can perform similar epistemic functions as highly idealized \textit{toy models}  \cite{SullivanForthcoming-SULDMM}. This indicates that a lack of representation based on similarity does not automatically deny the epistemic value of ML models.
}. This point is particularly evident in disciplines such as biology and economics, where researchers frequently employ highly idealized models that neither accurately describe nor predict the behavior of actual systems.

For instance, the Hawk-Dove game model in evolutionary biology (Maynard Smith and Price 1973  \cite{Smith1973TheLO}) is highly simplified relative to actual animal conflicts. The model assumes an ideal environment in which individuals possess equal fighting abilities and have no memory of previous encounters. It is not intended to provide an accurate description of any particular animal species. Rather, its purpose is to examine whether the evolution and maintenance of non-aggressive behavioral strategies are possible under individual selection alone, without invoking group selection (cf. Wirling and Grüne-Yanoff 2025  \cite{SjolinWirling2025-SJLTTP}). Cases such as this suggest that the epistemic value of scientific models cannot always be understood in purely representational terms. More recently, similar considerations have motivated attempts to evaluate the epistemic significance of machine learning models, whose internal structures are often highly complex and opaque (e.g., Sullivan 2022  \cite{Sullivan2022-SULUFM}; As Subotić 2024 \cite{Subotic2024-SUBEEB};  Verreault-Julien 2025  \cite{VerreaultJulien}).

To characterize the epistemic contribution of such models, philosophers of science frequently appeal to the notion of \textbf{how-possibly explanations} (HPEs). Although HPEs have various definitions (cf. Bokulich 2014  \cite{Bokulich2014-BOKHTT}; Brainard 2020 \cite{Brainard2020-BRAHTE-2}), it broadly refers to a form of explanation that shows how a phenomenon could occur in principle. This contrasts with \textbf{how-actually explanations} (HAEs), which aims to identify the actual mechanism responsible for the phenomenon. The distinction between HPEs and HAEs provides a useful framework for assessing the epistemic value of LLMs in theoretical linguistics. In the following sections, we first examine the explanatory contribution of LLMs when they are treated as minimal models that make no commitment to structural similarity or isomorphism with human linguistic cognition. We then consider the additional conditions that must be satisfied for LLMs to qualify as HAEs of human linguistic competence and language processing.

\section{The Minimal Epistemic Value of Language Models}

This section seeks to establish the epistemic value of LLMs under the most conservative assumptions, without presupposing that they are isomorphic to, or even closely resemble, human cognitive processes or neural structures. To this end, I draw on the framework of \textit{minimal models} developed in the philosophy of economics and argue that LLMs can function as tools for investigating modal claims in linguistics, namely claims concerning what is possible or necessary.

\subsection{Minimal Models and Their Epistemic Value}

Grüne-Yanoff (2009) characterizes minimal model as a model that ``lacks any similarity, isomorphism or resemblance relation to the world, to be unconstrained by natural laws or structural identity, and not to isolate any real factors'' (Grüne-Yanoff 2009: 83  \cite{Grune-Yanoff2009-GRNLFM-2}).

The notion of minimal models was originally introduced to account for common modeling practices in economics. Economic models are often highly idealized and are frequently constructed without the aim of accurately representing any particular real-world system. As Grüne-Yanoff (2009: 84) says,  ``no further claims are made about the truth of its assumptions, the epistemic status of the principles used in its construction, or the similarity of its economic interpretation (or parts of it) with the real world''. Despite this, economists construct and manipulate such models to learn about the real world. Similar practices can also be found in fields such as biology and physics (Pincock 2024  \cite{inbook}).

The question, then, is what kind of knowledge can be obtained from models that neither resemble reality nor accurately represent its causal structure. Grüne-Yanoff (2009: 85) defines learning from models as  ``a justified change in confidence in certain hypotheses about the world''. According to him, minimal models specifically change our degree of confidence in beliefs such as  ``phenomenon Y is impossible without  X'' or  `` X is necessary for Y to occur''. By presenting a credible situation where Y occurs without X, minimal models provide counterexamples to these beliefs and lower our commitment to them.

For learning to be possible, the scenario depicted by the minimal model must have credibility (Sugden 2000  \cite{Sugden01012000}). Credibility here does not mean the model is true or similar to reality. Rather, it means the scenario is convincing as a possible state of the world because it  ``coheres with common intuitions and experience'' (Sugden 2000: 26). As with a realistic novel, a model may describe a counterfactual world while remaining persuasive if the actions and processes it depicts appear internally consistent and intuitively plausible. Building on this idea, Grüne-Yanoff and Sjölin Wirling (2025) argue that a model supports justified modal inferences only when its scenario is internally coherent and judged plausible by competent users with appropriate background knowledge\cite{SjolinWirling2025-SJLTTP}. Consequently, assessments of credibility are always constrained by background knowledge, such as natural laws, knowledge of material components, and theories about relevant phenomena. Since evaluating credibility requires field-specific knowledge, these judgments can be relative to the discipline. 

As an example of learning from minimal models, Grüne-Yanoff (2009) points to Schelling's segregation model (Schelling 1971  \cite{Schelling01071971}). This model was created to explain the phenomenon of racial and ethnic segregation in human society. Schelling's model places two types of agents (like symbols or coins) on a grid. They move based on a simple rule: they prefer a certain percentage (e.g., one-third) of their neighbors to be of the same type. Although the model calls the symbols  \textit{neighbors}, their patterns  \textit{neighborhoods}, and gives them  \textit{preferences}, Schelling made no effort to justify these labels by pointing to similarities with actual cities, making the labels merely declared. Nevertheless, the model undermined people's belief that residential segregation is necessarily the result of strong racist preferences or systemic discrimination. Its epistemic contribution lies not in showing how segregation actually occurs, but in challenging the belief that strong discriminatory preferences are necessary for segregation to arise.

Minimal models therefore do not provide explanations of the actual causal mechanisms responsible for real-world phenomena (Fumagalli 2016  \cite{Fumagalli2016-FUMWWC}; cf. Pincock 2024  \cite{inbook}). Instead, they contribute by identifying objective possibilities and exploring alternative mechanisms through which a phenomenon could occur. In doing so, they revise our beliefs about what is possible, impossible, necessary, or contingent. Therefore, even if such models do not faithfully correspond to reality, they make an \textit{sui generis} epistemic contribution (Sjölin Wirling and Grüne-Yanoff 2024  \cite{SjolinWirling2024-SJLEAO}).

\subsection{Language Models as Minimal Models}

We now examine the epistemic value of LLMs under the most conservative assumption: the premise that there is absolutely no guarantee that their internal structures correspond to human cognitive mechanisms. Under this assumption, LLMs cannot be treated as explanatory models of actual human linguistic competence. Nevertheless, they may still possess epistemic value as minimal models.

Many influential claims in theoretical linguistics are modal in character. Generative approaches, for example, often maintain that certain properties of language are necessary for acquisition or linguistic competence. For example, the current Minimalist Program argues that syntactic operations like \textit{copy} and \textit{merge} are  ``conceptually necessary'' for language design (e.g., Chomsky 1995  \cite{chomsky1995minimalist}; Aoun et al. 2001  \cite{1360292619223300480}). Additionally, the Poverty of the Stimulus argument claims that because the linguistic input children receive is finite and incomplete, it is impossible for them to acquire complex grammatical knowledge without strong innate linguistic biases. These claims concern not merely how language actually is, but also how it could or could not be. From the perspective of modal modeling, LLMs can contribute to evaluating such claims even if they bear little resemblance to human cognitive mechanisms. If a language model successfully acquires a linguistic pattern under conditions that lack the theoretically posited constraint, this may function as a counterexample to the claim that the constraint is strictly necessary.

This perspective is closely related to \textit{the Proxy View} of LLMs (Ziv et al. 2025  \cite{ziv2026largelanguagemodelsproxies}; Johnsen 2025  \cite{johnsen2025grammaticalityjudgmentshumanslanguage}). According to this approach, LLMs should not be regarded as theories of human linguistic cognition. Rather, they serve as experimental tools for investigating what is learnable from linguistic experience alone, without explicit instruction or innate bias. Importantly, this use of LLMs does not presuppose any strong correspondences between model architecture and human cognition. In this respect, the Proxy View treats LLMs in a manner analogous to minimal models discussed in the philosophy of science.

Empirical findings within the Proxy View remain mixed. Ziv et al. (2025) report that many language models perform poorly on phenomena such as across-the-board (ATB) movement, parasitic gaps, and that-trace effects, and that they often learn typologically unattested language patterns more readily than natural languages. These findings suggest that current LLMs provide limited grounds for rejecting theories that posit strong innate linguistic biases. By contrast, Johnsen (2025) analyzed acceptability judgments on phenomena such as subject-auxiliary inversion and parasitic gaps using instruction-tuned language models, and found that the models were overall capable of judgments reflecting hierarchical syntactic structures. From this, he suggests that syntactic structures can be acquired through predictive training on surface forms, without explicit grammatical rules or innate knowledge. Despite these divergent findings, the Proxy View offers a coherent framework for using LLMs in linguistic inquiry. On this view, LLMs function as tools for evaluating and revising modal claims about language, particularly claims concerning what is necessary or impossible for language acquisition and linguistic competence. In this respect, they play a role analogous to that of minimal models in economics.

However, the epistemic value of such evidence depends crucially on the credibility of the scenario represented by the model. As discussed in Section 3.1, a minimal model can justify modal inferences only if the possibility it depicts is regarded as credible in light of relevant background knowledge. Consequently, the significance of LLM-based evidence depends not only on model performance but also on the background theories used to interpret that performance. These include both theories of language (e.g., generative grammar or usage-based approaches) and linking hypotheses connecting model behavior to human linguistic phenomena  (cf. Shah and Varma 2025). For example, claims relating model surprisal to human reading times or interpreting internal circuits as grammatical operations play a crucial role in determining what a model's behavior can legitimately be taken to demonstrate
\footnote{Furthermore, convergence across multiple language models with diverse architectures and training datasets can serve as a metric for credibility. When models with different designs consistently exhibit the same behavior toward a given linguistic phenomenon, the scenario they depict is less likely to be a mere artifact of any particular model, and its credibility as an HPE of human language is correspondingly strengthened.}. Therefore, LLMs can function as minimal models capable of supporting HPEs of language acquisition and linguistic competence. Yet the extent of their epistemic contribution depends on whether the scenarios they generate are judged credible under explicitly stated background theories and linking assumptions .

\section{Conditions for an LLM to Produce How-actually Explanations}
\subsection{Mechanistic Account of Explanation}

In scientific inquiry, accurate prediction alone does not constitute a scientific explanation. Scientific models can serve a variety of epistemic functions, including data summarization, prediction, and heuristic guidance. Not all models provide explanations for phenomena, even when they are highly useful. Craver (2006) distinguishes between \textit{phenomenal models} and \textit{explanatory models}. Phenomenal models characterize reliable relationships between inputs and outputs without specifying the mechanisms responsible for generating those relationships. Explanatory models, by contrast, identify the entities, activities, and organizational structures that produce the phenomenon of interest. \cite{Craver2006-CRAWMM}. For instance, Ptolemy’s geocentric model successfully predicted planetary motions but did not explain the physical causes underlying them. Similarly, the Hodgkin–Huxley model accurately described changes in neuronal membrane potential while leaving the underlying mechanisms of ion conductance unspecified. Such models therefore provide powerful descriptions and predictions without offering mechanistic explanations. In contrast, an explanatory model shows why a phenomenon occurs by revealing the underlying parts, entities, their activities, and their organized interactions.

What is required for models to go beyond a phenomenal model and become an explanatory one? Craver and Kaplan (2011) proposed \textbf{the model-to-mechanism-mapping (3M)} constraint for models to have explanatory force \cite{Kaplan}. This requires that some or all of the variables and dependencies in the model correspond to the actual components, activities, organization, and causal structures in the real target system. Under this condition, a phenomenal model that only describes input and output relationships is not a valid explanatory model. To possess explanatory force, models must describe not only these relationships but also the underlying mechanisms to some extent.

However, the original 3M constraint may be too restrictive for contemporary computational neuroscience because it appears to require relatively direct correspondence between model components and those of the target mechanism. To address this concern, Cao and Yamins (2024) proposed the \textbf{3M++ constraint}, which explicitly incorporates the role of abstraction in scientific models.\cite{CAO2024101244}. It consists of two specific requirements. The first is \textit{Predictively Adequate Runnable Abstraction} (PARA): A model must receive the same type of inputs as the target system, perform tasks relevant to the capacity under investigation, and successfully predict behavioral patterns exhibited by the target system. PARA ensures that the chosen level of abstraction preserves the causally relevant features necessary to generate the phenomenon. The second requirement is \textit{Transform Similarity}: The correspondence between a model and its target system must be established through the same kinds of transformations used to relate biological variation across individuals (such as the linear transform). Under the 3M++ constraint, the model’s internal structure does not need to be perfectly identical to the target system. Some physical details can be abstracted away as long as causally sufficient features are retained to produce the relevant capacity. 

As an example satisfying the 3M++ constraint, they cite the Deep Hierarchical Convo-lutional Neural Network (HCNN) as a model of the primate ventral visual pathway (Yamins et al. 2014 \cite{yaminsetal}). The HCNN is a multilayer neural network designed to mirror the hierarchical organization of visual processing, from early visual areas to the inferotemporal cortex (IT), and is optimized for object recognition. From the perspective of PARA, the HCNN satisfies both runnability and predictive adequacy. It receives the same type of input as humans and nonhuman primates (natural images) and performs an ecologically relevant task (object categorization). Moreover, its behavioral error patterns closely resemble those observed in humans and other primates performing comparable tasks. The model also satisfies Transform Similarity. Neural activity in different HCNN layers can be mapped onto neural responses recorded at different stages of the ventral visual pathway. In particular, higher layers best predict activity in the inferotemporal cortex, intermediate layers correspond most closely to V4, and lower layers align with V1. This provides evidence that a true mechanistic correspondence exists between the HCNN and the ventral visual pathway.

The discussion above suggests that the 3M++ constraint therefore provides a useful criterion for determining whether a model merely offers a HPE or genuinely qualifies as a HAE of the target system. In the following section, we apply this framework to contemporary LLMs and assess whether they meet the conditions necessary for mechanistic explanation of human linguistic cognition.

\subsection{Applying to LLMs}

If we apply the mechanistic framework discussed in the previous section to current LLMs, what conditions must they satisfy in order to qualify as HAEs of human language capacity?

\paragraph{Predictively Adequate Runnable Abstraction}

According to Predictively Adequate Runnable Abstraction (PARA), the first requirement of the 3M++ constraint, models must receive the same type of input as the target system and actually perform tasks related to the capacity of interest. LLMs partially satisfy this condition in specific domains.  As Mahowald et al. (2023) point out, LLMs achieve a considerable level of formal linguistic competence, which is the ability to correctly handle grammatical forms  \cite{MAHOWALD2024517}. For example, in syntactic minimal pair benchmarks like BLiMP (Warstadt et al. 2020  \cite{warstadt-etal-2020-blimp-benchmark}), models such as GPT-2 show high accuracy rates on phenomena like subject-verb agreement. This indicates that LLMs can capture complex syntactic patterns to some extent. However, when evaluating PARA, the question is what to consider as the task of interest. Depending on the linguist's focus, different behavioral metrics could be targeted, such as acceptability judgments on benchmarks and reading times.

For instance, in surprisal theory regarding human sentence processing (e.g., Hale 2001  \cite{10.3115/1073336.1073357}), the correlation between surprisal in information theory and human reading times is a key issue. However, Huang et al. (2024) showed that LLM surprisal systematically underpredicts the difficulty of processing syntactic ambiguities \cite{HUANG2024104510}. In particular, it fails to predict processing difficulty in syntactically complex sentences like garden-path sentences. This shows that high benchmark accuracy does not imply that LLMs broadly predict the behavioral patterns observed in human language processing.

From the perspective of language acquisition, PARA would require LLMs to receive the same quality and quantity of linguistic input as children. Current standard LLMs do not meet this condition. While children acquire their native language from fewer than 100 million words of input by around age 13, LLMs are typically trained on thousands to hundreds of thousands of times more tokens (Warstadt et al. 2023a  \cite{warstadt2023papersbabylmchallenge}). The BabyLM Challenge (Warstadt et al. 2023b  \cite{Warstadt_2023}) directly addresses this issue. This project aims to build data-efficient language models and explore cognitively plausible computational models by training LMs on corpora restricted to 100 million words or less, similar to human child input. Through architectural improvements and adjustments to training epochs, these models demonstrated performance comparable to Llama 2, other massive models trained on trillions of words, and human scores in metrics like grammatical knowledge and language understanding  \cite{Warstadt_2023}. This suggests that robust linguistic generalization can be learned from human-scale data. However, many top models relied on repeated training over hundreds to thousands of epochs to achieve high accuracy from limited data. While they successfully restricted the data volume, they consumed massive computational resources instead. Furthermore, methods inspired by human learning, such as multimodal learning, did not always prove effective. Therefore, reservations remain about whether these models truly possess cognitive and developmental plausibility.

\paragraph{Transform Similarity}

Next, Transform Similarity requires that the mapping between the model and the target system be constructed using the same type of transformations  used to match individuals with biological variations. Recently, inspired by the success of Deep Convolutional Neural Networks (DCNNs) in vision research, there have been attempts to apply similar methods to the language domain.  For example, Schrimpf et al. (2021) tested a diverse set of language models using fMRI and ECoG data to see how accurately they could predict human neural activity  \cite{schrimpf2021}. They found that Transformer-based architectures predicted a substantial proportion of explainable variance in neural responses to sentences. Notably, even untrained Transformer models achieved nontrivial predictive performance, suggesting that architectural properties alone may capture aspects of neural organization relevant to language.

However, the interpretation of these findings remains controversial. Reanalyzing the same fMRI dataset, Feghhi et al. (2024) argue that much of the observed predictive success can be accounted for by relatively simple variables, including sentence length and word position\cite{feghhi2024largelanguagemodelsmapping}.
According to their analysis, contextual syntactic representations contribute only limited additional explanatory power. If correct, these findings undermine the claim that current brain-score results reveal a meaningful mechanistic correspondence between language models and human neural systems. Rather, they may actually just be reacting to theoretically uniformative features. Furthermore, compared to visual processing, how human language processing occurs is anatomically and functionally uncertain. In visual processing, the ventral visual pathway has a clear hierarchical structure (V1, V4, and IT), making it relatively clear which stage corresponded to each layer of the HCNN. In contrast, the brain mechanisms for language processing are far more ambiguous. For example, Tremblay and Dick (2016) surveyed language neuroscientists on the anatomical definitions of basic language areas like Broca’s and Wernicke’s areas \cite{Tremblay2016BrocaAW}. They revealed that there are no consistent definitions within the research community. Functionally as well, human language processing is not confined to specific local areas. According to Fedorenko et al. (2024), language processing functions as a distributed network across the frontal and temporal lobes \cite{Fedorenko2024TheLN}. Additionally, Blank and Fedorenko (2020) showed that Broca’s area, traditionally considered the core of language processing, is not a functionally homogeneous single region. Instead, it is a collection of parts involved in language processing and multiple-demand parts involved in working memory, calculation, and attention control \cite{FEDORENKO2020270}. This anatomical and functional uncertainty creates difficulties in evaluating Transform Similarity. In the case of the ventral visual pathway, there was a reference hierarchical structure to verify the mapping between HCNN layers and brain regions. In language processing, the underlying brain processing stages are not sufficiently understood. Consequently, there is no established standard for what constitutes a successful mapping between a model and the brain. Therefore, even if LLMs appear to predict the brain’s language activity, there is currently no criterion to distinguish whether this shows a true mechanistic correspondence or merely a superficial correlation.

In summary, the assessment of LLMs through the lenses of both PARA and Transform Similarity suggests that it remains premature to conclude that they satisfy the 3M++ constraint. With respect to PARA, although contemporary LLMs achieve impressive performance on formal linguistic competence, they do not yet reliably predict a broad range of human language-processing behaviors, such as reading-time patterns. Furthermore, from the perspective of language acquisition, current models differ substantially from human learners in both the quantity and quality of the linguistic input they receive. Their failure to satisfy PARA may indicate that they do not yet preserve the causally relevant features underlying those phenomena. With respect to Transform Similarity, the neural mechanisms underlying language processing remain considerably less well understood than those underlying vision. As a result, it is currently difficult to determine whether observed correspondences between language models and neural activity reflect genuine mechanistic similarities or not. Nevertheless, this does not imply that LLMs are incapable, in principle, of providing HPEs of human linguistic cognition. As the BabyLM Challenge shows, models are being continuously improved in terms of cognitive and developmental plausibility. Elucidating the neural basis of language processing also remains an ongoing research goal. As the neural basis of language processing becomes clearer and models improve in cognitive and developmental plausibility, we will be able to make a clearer judgment on whether LLMs satisfy the requirements of the 3M++ constraint.

\section{Implications of this Paper
}
Based on the discussions in the previous sections, this section clarifies where existing positions on the relationship between LLMs and linguistic theory are flawed or insufficient.

\subsection{Re-evaluating Existing Positions}

Insulationists argue that LLMs are irrelevant to theoretical linguistics because they rely on vastly more data than children receive during language acquisition, fail to exhibit key features of human linguistic cognition such as compositionality and modularity, and offer only probabilistic predictions rather than genuine scientific explanations. The insulationist critique rests on a hidden assumption that the only scientifically legitimate models are those that describe actual causal mechanisms and maintain structural correspondence with their target system. However, as argued in Section 3, this assumption overlooks the epistemic contribution that models can make as HPEs. Even if LLMs do not mirror the actual mechanisms underlying human language acquisition or processing, they can still provide credible demonstrations that certain linguistic capacities may emerge without assumptions that have often been regarded as necessary, such as strong language-specific innate constraints. In this respect, LLMs can function as minimal models that explore the space of possible mechanisms for language learning and linguistic competence. However, by dismissing LLMs solely on the grounds that they fall short of HAEs, insulationists fail to consider whether LLMs might contribute at the level of HPEs. 

Eliminativists err in the opposite direction. Piantadosi (2024) and others argue that because LLMs acquire sophisticated syntactic and semantic knowledge through next-token prediction alone, without explicit grammatical rules or innate universal grammar, this constitutes a refutation of the Chomskyan approach and grounds for replacing traditional linguistic theories with LLMs themselves. This position, however, conflates HPEs with HAEs. The fact that an LLM successfully reproduces human-like linguistic behavior demonstrates that a particular route to linguistic competence is possible. But it does not establish that the same route is the one actually employed by humans. Eliminativists overestimate the success of LLMs as HPEs, which merely present possibilities, by treating them as successes as HAEs that elucidate actual target mechanisms. 

Conciliationists propose that LLMs can serve as tools for constraining and testing hypotheses about language acquisition and competence, without fully replacing existing theories. They correctly avoid the all-or-nothing view of explanatory value of LLMs shared by insulationists and eliminativists. Nevertheless, it often remains unclear exactly what kind of epistemic contribution LLMs make and why such contributions are justified. For example, Futrell and Mahowald (2025) suggest that the ability of LLMs to acquire gradient and probabilistic knowledge of grammatical constructions provides a proof of concept for usage-based approaches \cite{Futrell_2025}. Yet the precise epistemic significance of such a proof of concept remains unclear.

\subsection{Toward a Gradient View of Epistemic Value of Language Models}

The framework developed in this paper suggests that the distinction between HPEs and HAEs should not be understood as a sharp boundary. Although these categories mark two different explanatory ideals, the epistemic value of scientific models often develops gradually. There is increasing evidence that LLMs capture linguistically relevant structure beyond surface-level statistical regularities. For example, Hewitt and Manning (2019) demonstrated using a structural probe that syntactic distance and depth information is embedded in the vector space of word representations in models such as BERT and ELMo \cite{hewitt-manning-2019-structural}. Ahuja et al. (2025) showed that Transformer-based models exhibit human-like generalization to non-local subject–verb agreement even when such constructions are absent from the training data \cite{ahuja-etal-2025-learning}. These findings suggest that the epistemic value of LLMs is best understood as occupying different positions along a continuum between HPEs and HAEs. At one end of this continuum are models that function primarily as HPEs. Such models provide credible counterexamples to claims of necessity or impossibility while making no commitment to describing the actual mechanisms underlying human language. At the other end are models that satisfy the requirements of mechanistic explanation and therefore qualify as HAEs. Between these poles lie a variety of intermediate cases in which models exhibit varying degrees of explanatory relevance to human linguistic cognition.

This continuum view helps clarify the status of several current research programs. The proxy view discussed in Section 3 is located close to the HPE end of the spectrum. Its primary contribution is to explore whether particular linguistic capacities can emerge under specified learning conditions, thereby informing modal claims about language acquisition. Research on developmentally plausible training regimes, such as the BabyLM initiative, can be understood as moving toward the HAE end of the spectrum by reducing the gap between model learning conditions and those of human learners. Similarly, studies that relate language-model representations to neural activity aim to identify potential correspondences between model internals and human cognitive mechanisms. Although current evidence remains insufficient to establish Transform Similarity in the sense required by the 3M++ framework, such work can nevertheless be understood as narrowing the explanatory gap between language models and human cognition. Viewing LLMs through this lens also helps clarify the epistemic status of conciliationist proposals. When Futrell and Mahowald (2025) describe LLM experiments as a proof of concept for usage-based approaches, the contribution is best understood at the level of HPEs. Such experiments can provide credible scenarios that challenge claims about what is necessary for language acquisition. For example, they may support the possibility that typologically natural linguistic patterns can emerge without language-specific innate constraints. In this sense, an LLM-based proof of concept can motivate revisions to modal beliefs about human language rather than as explanations of what actually occurs. 

The practical implication is that discussions of LLMs in linguistics should focus less on whether a model succeeds or fails as an explanation in absolute terms and more on where it is situated along the continuum between HPEs and HAEs. Different explanatory aims call for different standards of evaluation, and models should therefore be assessed relative to the epistemic role they are intended to play. This perspective allows for a more nuanced comparison of research programs pursuing different explanatory goals while avoiding the tendency either to dismiss or to overstate the scientific significance of language models.

\section{Conclusion}

We have argued that the epistemic significance of LLMs for theoretical linguistics should not be evaluated in all-or-nothing terms. Drawing on the framework of modal modeling, we distinguished between HPEs and HAEs and used this distinction to clarify the different ways in which LLMs can contribute to linguistic inquiry. We proposed a more nuanced, gradient view according to which LLMs may occupy different positions along a continuum extending from HPEs to HAEs. On this view, LLMs can possess genuine epistemic value even when treated as minimal models that lack any established structural correspondence with human cognitive mechanisms. Their contribution lies in their capacity to function as HPEs, providing credible scenarios that support or challenge modal claims about language acquisition and linguistic competence. At the same time, we argued that providing HAEs of human linguistic capacity requires satisfying the 3M++ conditions, including PARA and Transform Similarity. Current LLMs do not yet fully meet these requirements. To advance the debate, we propose that researchers make three questions explicit when employing LLMs as evidence in linguistic research. First, what epistemic role is the model intended to play? Second, if the model is used primarily as an HPE, which modal claim is it intended to support or challenge? Third, if the model is intended to move toward an HAE, what aspect of human linguistic capacity is being explained, and at what level of abstraction is the explanation being formulated?

By explicitly addressing these questions, researchers can more clearly articulate the epistemic aims of their work and the standards by which their claims should be evaluated. We hope that this framework will provide a common basis for assessing the epistemic significance of LLMs and foster more productive dialogue across competing theoretical traditions in linguistics.

\bibliographystyle{plain}
\bibliography{Language}

@article{ahuja-etal-2025-learning,
    title = "Learning Syntax Without Planting Trees: Understanding Hierarchical Generalization in Transformers",
    author = "Ahuja, Kabir  and
      Balachandran, Vidhisha  and
      Panwar, Madhur  and
      He, Tianxing  and
      Smith, Noah A.  and
      Goyal, Navin  and
      Tsvetkov, Yulia",
    journal = "Transactions of the Association for Computational Linguistics",
    volume = "13",
    year = "2025",
    address = "Cambridge, MA",
    publisher = "MIT Press",
    url = "https://aclanthology.org/2025.tacl-1.6/",
    doi = "10.1162/tacl_a_00733",
    pages = "121--141"
}

@article{CAO2024101244,
title = {Explanatory models in neuroscience, Part 1: Taking mechanistic abstraction seriously},
journal = {Cognitive Systems Research},
volume = {87},
pages = {101244},
year = {2024},
issn = {1389-0417},
doi = {https://doi.org/10.1016/j.cogsys.2024.101244},
url = {https://www.sciencedirect.com/science/article/pii/S138904172400038X},
author = {Rosa Cao and Daniel Yamins},
}

@article{A&B2024,
author = {Ambridge, Ben and Blything, Liam},
year = {2024},
month = {07},
pages = {33-48},
title = {Large language models are better than theoretical linguists at theoretical linguistics},
volume = {50},
journal = {Theoretical Linguistics},
doi = {10.1515/tl-2024-2002}
}

@article{1360292619223300480,
author="Joseph, Aoun and Lina, Choueiri and Norbert, Hornstein",
title="Resumption, Movement, and Derivational Economy",
journal="Linguistic Inquiry",
ISSN="0024-3892",
publisher="MIT Press - Journals",
year="2001",
month="07",
volume="32",
number="3",
pages="371-403",
DOI="10.1162/002438901750372504",
URL="https://cir.nii.ac.jp/crid/1360292619223300480"
}

@article{FEDORENKO2020270,
title = {Broca’s Area Is Not a Natural Kind},
journal = {Trends in Cognitive Sciences},
volume = {24},
number = {4},
pages = {270-284},
year = {2020},
issn = {1364-6613},
doi = {https://doi.org/10.1016/j.tics.2020.01.001},
url = {https://www.sciencedirect.com/science/article/pii/S1364661320300036},
author = {Evelina Fedorenko and Idan A. Blank}
}

@article{Bokulich2014-BOKHTT,
	author = {Alisa Bokulich},
	doi = {10.5840/monist201497321},
	journal = {The Monist},
	number = {3},
	pages = {321--338},
	publisher = {The Hegeler Institute},
	title = {How the Tiger Bush Got its Stripes: ?How Possibly? Vs. ?How Actually?Model Explanations},
	volume = {97},
	year = {2014}
}

@misc{bolhuis2025largelanguagemodelsnatural,
      title={Large language models are not about natural language}, 
      author={Johan J. Bolhuis and Andrea Moro and Stephen Crain and Sandiway Fong},
      year={2025},
      eprint={2512.13441},
      archivePrefix={arXiv},
      primaryClass={cs.CL},
      url={https://arxiv.org/abs/2512.13441}, 
}

@article{Brainard2020-BRAHTE-2,
	author = {Lindsay Brainard},
	journal = {Philosophers' Imprint},
	number = {13},
	pages = {1--23},
	title = {How to Explain How-Possibly},
	volume = {20},
	year = {2020}
}

@book{chomsky1995minimalist,
  address = {Cambridge, MA},
  author = {Chomsky, N},
  publisher = {MIT Press},
  title = {The Minimalist Program},
  year = 1995
}

@article{Chomskyetal2023,
author = {Chomsky, N. Roberts, I. \& J. Watumull },
year = {2023},
month = {05},
pages = {},
title = {Noam Chomsky: The False Promise of ChatGPT},
journal = {The New York Times}
}

@article{Craver2006-CRAWMM,
	author = {Carl F. Craver},
	doi = {10.1007/s11229-006-9097-x},
	journal = {Synthese},
	number = {3},
	pages = {355--376},
	publisher = {Springer},
	title = {When Mechanistic Models Explain},
	volume = {153},
	year = {2006}
}

@article{Dupre2021-DUPWCD,
	author = {Gabe Dupre},
	doi = {10.1007/s11023-021-09571-w},
	journal = {Minds and Machines},
	number = {4},
	pages = {617--635},
	publisher = {Springer Verlag},
	title = {(What) Can Deep Learning Contribute to Theoretical Linguistics?},
	volume = {31},
	year = {2021}
}

@article{Fedorenko2024TheLN,
  title={The language network as a natural kind within the broader landscape of the human brain},
  author={Evelina Fedorenko and Anna A. Ivanova and Tamar I. Regev},
  journal={Nature Reviews Neuroscience},
  year={2024},
  volume={25},
  pages={289 - 312},
  url={https://api.semanticscholar.org/CorpusID:269112224}
}

@misc{feghhi2024largelanguagemodelsmapping,
      title={What Are Large Language Models Mapping to in the Brain? A Case Against Over-Reliance on Brain Scores}, 
      author={Ebrahim Feghhi and Nima Hadidi and Bryan Song and Idan A. Blank and Jonathan C. Kao},
      year={2024},
      eprint={2406.01538},
      archivePrefix={arXiv},
      primaryClass={cs.CL},
      url={https://arxiv.org/abs/2406.01538}, 
}

@article{FoxandKazir,
author = {Fox, Danny and Katzir, Roni},
year = {2024},
month = {07},
pages = {71-76},
title = {Large Language Models and theoretical linguistics},
volume = {50},
journal = {Theoretical Linguistics},
doi = {10.1515/tl-2024-2005}
}

@article{Fumagalli2016-FUMWWC,
	author = {Roberto Fumagalli},
	doi = {10.1007/s10670-015-9749-7},
	journal = {Erkenntnis},
	number = {3},
	pages = {433--455},
	publisher = {Springer Nature},
	title = {Why We Cannot Learn From Minimal Models},
	volume = {81},
	year = {2016}
}

@article{Futrell_2025,
   title={How Linguistics Learned to Stop Worrying and Love the Language Models},
   ISSN={1469-1825},
   url={http://dx.doi.org/10.1017/S0140525X2510112X},
   DOI={10.1017/s0140525x2510112x},
   journal={Behavioral and Brain Sciences},
   publisher={Cambridge University Press (CUP)},
   author={Futrell, Richard and Mahowald, Kyle},
   year={2025},
   month=jul, pages={1–98} }

@article{Giere2004-GIEHMA,
	author = {Ronald N. Giere},
	doi = {10.1086/425063},
	journal = {Philosophy of Science},
	number = {5},
	pages = {742--752},
	publisher = {University of Chicago Press},
	title = {How Models Are Used to Represent Reality},
	volume = {71},
	year = {2004}
}

@article{Grune-Yanoff2009-GRNLFM-2,
	author = {Till Gr\"{u}ne{-}Yanoff},
	doi = {10.1007/s10670-008-9138-6},
	journal = {Erkenntnis},
	number = {1},
	pages = {81--99},
	title = {Learning From Minimal Economic Models},
	volume = {70},
	year = {2009}
}

@inproceedings{10.3115/1073336.1073357,
author = {Hale, John},
title = {A probabilistic earley parser as a psycholinguistic model},
year = {2001},
publisher = {Association for Computational Linguistics},
address = {USA},
url = {https://doi.org/10.3115/1073336.1073357},
doi = {10.3115/1073336.1073357}}

@inproceedings{hewitt-manning-2019-structural,
    title = "{A} Structural Probe for Finding Syntax in Word Representations",
    author = "Hewitt, John  and
      Manning, Christopher D.",
    editor = "Burstein, Jill  and
      Doran, Christy  and
      Solorio, Thamar",
    booktitle = "Proceedings of the 2019 Conference of the North {A}merican Chapter of the Association for Computational Linguistics: Human Language Technologies, Volume 1 (Long and Short Papers)",
    month = jun,
    year = "2019",
    address = "Minneapolis, Minnesota",
    publisher = "Association for Computational Linguistics",
    url = "https://aclanthology.org/N19-1419/",
    doi = "10.18653/v1/N19-1419",
    pages = "4129--4138",}

@article{HUANG2024104510,
title = {Large-scale benchmark yields no evidence that language model surprisal explains syntactic disambiguation difficulty},
journal = {Journal of Memory and Language},
author = {Kuan-Jung Huang and Suhas Arehalli and Mari Kugemoto and Christian Muxica and Grusha Prasad and Brian Dillon and Tal Linzen},
volume = {137},
pages = {104510},
year = {2024},
issn = {0749-596X},
doi = {https://doi.org/10.1016/j.jml.2024.104510},
url = {https://www.sciencedirect.com/science/article/pii/S0749596X24000135}
}

@misc{johnsen2025grammaticalityjudgmentshumanslanguage,
      title={Grammaticality Judgments in Humans and Language Models: Revisiting Generative Grammar with LLMs}, 
      author={Lars G. B. Johnsen},
      year={2025},
      eprint={2512.10453},
      archivePrefix={arXiv},
      primaryClass={cs.CL},
      url={https://arxiv.org/abs/2512.10453}, 
}

@article{Kaplan,
author = {Kaplan, David and Craver, Carl},
year = {2011},
month = {11},
pages = {601-627},
title = {The Explanatory Force of Dynamical and Mathematical Models in Neuroscience: A Mechanistic Perspective*},
volume = {78},
journal = {Philosophy of Science},
doi = {10.1086/661755}
}

@book{book,
author = {Knuuttila, Tarja and Grüne-Yanoff, Till and Koskinen, Rami and Wirling, Ylwa},
year = {2025},
month = {02},
pages = {},
title = {Modeling the Possible: Perspectives from Philosophy of Science},
isbn = {9781003342816},
doi = {10.4324/9781003342816}
}

@misc{kodner2023linguisticsthrive21stcentury,
      title={Why Linguistics Will Thrive in the 21st Century: A Reply to Piantadosi (2023)}, 
      author={Jordan Kodner and Sarah Payne and Jeffrey Heinz},
      year={2023},
      eprint={2308.03228},
      archivePrefix={arXiv},
      primaryClass={cs.CL},
      url={https://arxiv.org/abs/2308.03228}, 
}

@article{MAHOWALD2024517,
title = {Dissociating language and thought in large language models},
journal = {Trends in Cognitive Sciences},
author={Kyle Mahowald and Anna A. Ivanova and Idan A. Blank and Nancy Kanwisher and Joshua B. Tenenbaum and Evelina Fedorenko},
volume = {28},
number = {6},
pages = {517-540},
year = {2024},
issn = {1364-6613},
doi = {https://doi.org/10.1016/j.tics.2024.01.011},
url = {https://www.sciencedirect.com/science/article/pii/S1364661324000275}
}

@article{Smith1973TheLO,
  title={The Logic of Animal Conflict},
  author={J. Maynard Smith and G. Randall Price},
  journal={Nature},
  year={1973},
  volume={246},
  pages={15-18},
  url={https://api.semanticscholar.org/CorpusID:4224989}
}

@article{doi:10.1177/09637214241268098,
author = {Sam Whitman McGrath and Jacob Russin and Ellie Pavlick and Roman Feiman},
title ={How Can Deep Neural Networks Inform Theory in Psychological Science?},

journal = {Current Directions in Psychological Science},
volume = {33},
number = {5},
pages = {325-333},
year = {2024},
doi = {10.1177/09637214241268098},
URL = { 
        https://doi.org/10.1177/09637214241268098
},
eprint = { 
        https://doi.org/10.1177/09637214241268098
}
}

@article{milliere,


author = {Millière, Raphaël},
year = {2024},
month = {08},
pages = {},
title = {Language Models as Models of Language},
doi = {10.48550/arXiv.2408.07144}
}

@article{MORO202382,
title = {Large languages, impossible languages and human brains},
journal = {Cortex},
volume = {167},
pages = {82-85},
year = {2023},
issn = {0010-9452},
doi = {https://doi.org/10.1016/j.cortex.2023.07.003},
url = {https://www.sciencedirect.com/science/article/pii/S0010945223001752},
author = {Andrea Moro and Matteo Greco and Stefano F. Cappa}
}

@article{articleMuller,
author = {Müller, Stefan},
year = {2025},
month = {04},
pages = {1–15},
title = {The best linguistic theory, a wrong linguistic theory, or no theory at all?},
volume = {44},
journal = {Zeitschrift für Sprachwissenschaft},
doi = {10.18148/zs/2025-2001}
}

@incollection{piantadosi2023modern,
  title={Modern language models refute Chomsky's approach to language},
  author={Steven T. Piantadosi},
  year={2024},
  booktitle={From fieldwork to linguistic theory: {A tribute to Dan Everett (Empirically
Oriented Theoretical Morphology and Syntax 15)}},
  editor={Edward Gibson and Moshe Poliak},
  pages={353---414},
  publisher={Berlin: Language Science Press},
  url={https://lingbuzz.net/lingbuzz/007180}
}

@inbook{inbook,
author = {Pincock, Christopher},
year = {2024},
month = {08},
pages = {138-148},
title = {Minimal models},
isbn = {9781003205647},
doi = {10.4324/9781003205647-13}
}

@article{
schrimpf2021,
author = {Martin Schrimpf  and Idan Asher Blank  and Greta Tuckute  and Carina Kauf  and Eghbal A. Hosseini  and Nancy Kanwisher  and Joshua B. Tenenbaum  and Evelina Fedorenko },
title = {The neural architecture of language: Integrative modeling converges on predictive processing},
journal = {Proceedings of the National Academy of Sciences},
volume = {118},
number = {45},
pages = {e2105646118},
year = {2021},
doi = {10.1073/pnas.2105646118},
URL = {https://www.pnas.org/doi/abs/10.1073/pnas.2105646118},
eprint = {https://www.pnas.org/doi/pdf/10.1073/pnas.2105646118}}

@article{Schelling01071971,
author = {Thomas C. Schelling},
title = {Dynamic models of segregation† },
journal = {The Journal of Mathematical Sociology},
volume = {1},
number = {2},
pages = {143--186},
year = {1971},
publisher = {Routledge},
doi = {10.1080/0022250X.1971.9989794},
URL = {    
        https://doi.org/10.1080/0022250X.1971.9989794
},
eprint = { 
    
        https://doi.org/10.1080/0022250X.1971.9989794
}
}

@article{Subotic2024-SUBEEB,
	author = {Vanja Suboti\'c},
	doi = {10.1007/s11229-024-04514-1},
	journal = {Synthese},
	number = {3},
	pages = {1--28},
	publisher = {Springer Verlag},
	title = {Exploring, Expounding \& Ersatzing: A Three-Level Account of Deep Learning Models in Cognitive Neuroscience},
	volume = {203},
	year = {2024}
}

@article{Sugden01012000,
author = {Robert Sugden},
title = {Credible worlds: the status of theoretical models in economics},
journal = {Journal of Economic Methodology},
volume = {7},
number = {1},
pages = {1--31},
year = {2000},
publisher = {Routledge},
doi = {10.1080/135017800362220},
URL = { 
        https://doi.org/10.1080/135017800362220
},
eprint = {    
        https://doi.org/10.1080/135017800362220
}
}

@article{Sullivan2022-SULUFM,
	author = {Emily Sullivan},
	doi = {10.1093/bjps/axz035},
	journal = {British Journal for the Philosophy of Science},
	number = {1},
	pages = {109--133},
	publisher = {University of Chicago Press},
	title = {Understanding From Machine Learning Models},
	volume = {73},
	year = {2022}
}

@article{SullivanForthcoming-SULDMM,
	author = {Emily Sullivan},
	journal = {Philosophy of Science},
	title = {Do Ml Models Represent Their Targets?},
	year = {forthcoming}
}

@article{Tremblay2016BrocaAW,
  title={Broca and Wernicke are dead, or moving past the classic model of language neurobiology.},
  author={Pascale Tremblay and Anthony Steven Dick},
  journal={Brain and language},
  year={2016},
  volume={162},
  pages={
          60-71
        },
  url={https://api.semanticscholar.org/CorpusID:3344826}
}

@inbook{VerreaultJulien,
author = {Verreault-Julien, Philippe},
year = {2025},
month = {01},
pages = {177-195},
title = {Three Strategies for Salvaging Explanatory Value in Deep Neural Network Modeling},
isbn = {9781003342816},
doi = {10.4324/9781003342816-13}
}

@article{warstadt-etal-2020-blimp-benchmark,
    title = "{BL}i{MP}: The Benchmark of Linguistic Minimal Pairs for {E}nglish",
    author = "Warstadt, Alex  and
      Parrish, Alicia  and
      Liu, Haokun  and
      Mohananey, Anhad  and
      Peng, Wei  and
      Wang, Sheng-Fu  and
      Bowman, Samuel R.",
    editor = "Johnson, Mark  and
      Roark, Brian  and
      Nenkova, Ani",
    journal = "Transactions of the Association for Computational Linguistics",
    volume = "8",
    year = "2020",
    address = "Cambridge, MA",
    publisher = "MIT Press",
    url = "https://aclanthology.org/2020.tacl-1.25/",
    doi = "10.1162/tacl_a_00321",
    pages = "377--392"
}

@misc{warstadt2023papersbabylmchallenge,
      title={Call for Papers -- The BabyLM Challenge: Sample-efficient pretraining on a developmentally plausible corpus}, 
      author={Alex Warstadt and Leshem Choshen and Aaron Mueller and Adina Williams and Ethan Wilcox and Chengxu Zhuang},
      year={2023},
      eprint={2301.11796},
      archivePrefix={arXiv},
      primaryClass={cs.CL},
      url={https://arxiv.org/abs/2301.11796}, 
}

@inproceedings{Warstadt_2023,
   title={Findings of the BabyLM Challenge: Sample-Efficient Pretraining on Developmentally Plausible Corpora},
   url={http://dx.doi.org/10.18653/v1/2023.conll-babylm.1},
   DOI={10.18653/v1/2023.conll-babylm.1},
   booktitle={Proceedings of the BabyLM Challenge at the 27th Conference on Computational Natural Language Learning},
   publisher={Association for Computational Linguistics},
   author={Warstadt, Alex and Mueller, Aaron and Choshen, Leshem and Wilcox, Ethan and Zhuang, Chengxu and Ciro, Juan and Mosquera, Rafael and Paranjabe, Bhargavi and Williams, Adina and Linzen, Tal and Cotterell, Ryan},
   year={2023},
   pages={1–6} }

@book{Weisberg2013-WEISAS-5,
	address = {New York, US},
	author = {Michael Weisberg},
	editor = {},
	publisher = {Oxford University Press},
	title = {Simulation and Similarity: Using Models to Understand the World},
	year = {2013}
}

@incollection{SjolinWirling2025-SJLTTP,
	author = {Ylwa Sj\"{o}lin Wirling and Till Gr\"{u}ne{-}Yanoff},
	booktitle = {Modeling the Possible. Perspectives from Philosophy of Science},
	editor = {Tarja Knuuttila and Till Gr\"{u}ne{-}Yanoff and Rami Koskinen and Ylwa Wirling},
	pages = {27--47},
	publisher = {Routledge},
	title = {Through the Prism of Modal Epistemology: Perspective on Modal Modeling},
	year = {2025}
}

@article{SjolinWirling2024-SJLEAO,
	author = {Ylwa Sj\"{o}lin Wirling and Till Gr\"{u}ne{-}Yanoff},
	doi = {10.1086/716925},
	journal = {British Journal for the Philosophy of Science},
	number = {4},
	pages = {821--841},
	title = {Epistemic and Objective Possibility in Science},
	volume = {75},
	year = {2024}
}

@article{yaminsetal,
author = {Yamins, Daniel and Hong, Ha and Cadieu, Charles and Solomon, Ethan and Seibert, Darren and Dicarlo, James},
year = {2014},
month = {05},
pages = {},
title = {Performance-optimized hierarchical models predict neural responses in higher visual cortex},
volume = {111},
journal = {Proceedings of the National Academy of Sciences of the United States of America},
doi = {10.1073/pnas.1403112111}
}

@misc{ziv2026largelanguagemodelsproxies,
      title={Large Language Models as Proxies for Theories of Human Linguistic Cognition}, 
      author={Imry Ziv and Nur Lan and Emmanuel Chemla and Roni Katzir},
      year={2026},
      eprint={2502.07687},
      archivePrefix={arXiv},
      primaryClass={cs.CL},
      url={https://arxiv.org/abs/2502.07687}, 
}

@article{S&G2011,
author = {Sjölin Wirling, Ylwa and Grüne-Yanoff, Till},
title = {The epistemology of modal modeling},
journal = {Philosophy Compass},
year={2021},
volume = {16},
number = {10},
pages = {e12775},
doi = {https://doi.org/10.1111/phc3.12775}
}

\end{document}